\documentclass[10pt,twocolumn,letterpaper]{article}
\usepackage{cvpr}
\usepackage{times}
\usepackage{epsfig}
\usepackage{graphicx}
\usepackage{amsmath}
\usepackage{amssymb}
\usepackage{algorithm}
\usepackage{amsmath}
\usepackage{color}
\usepackage{graphviz}

\usepackage[numbers]{natbib}
\usepackage[breaklinks=true,bookmarks=false]{hyperref}
\cvprfinalcopy 

\def\cvprPaperID{4099} 

\usepackage{colortbl,bbm}

\definecolor{DarkGreen}   {rgb}{0,0.5,0}
\definecolor{DarkBlue}    {rgb}{0,0.0,0.5}
\definecolor{LightGray}   {rgb}{0.8,0.8,0.8}

\newcommand{\av}[0]{{\boldsymbol{a}}}

\newcommand{\xv}{\boldsymbol{x}}

\newcommand{\zv}{\boldsymbol{z}}

\newcommand{\Dcal}{\mathcal{D}}

\begin{document}

\title{Generalized Zero-Shot Learning via Synthesized Examples}

\author{Vinay Kumar Verma$^{1,}$\footnotemark[1] , Gundeep Arora$^{1,}$\footnotemark[1], Ashish Mishra$^\dag$ and Piyush Rai$^*$\\
$^*$Indian Institute of Technology Kanpur \qquad $^\dag$Indian Institute of Technology Madras\\
{\tt\small \{vkverma,gundeep,piyush\}@cse.iitk.ac.in, mishra@cse.iitm.ac.in}
}
\maketitle
\begin{abstract}
We present a generative framework for generalized zero-shot learning where the training and test classes are not necessarily disjoint. Built upon a variational autoencoder based architecture, consisting of a probabilistic encoder and a probabilistic \emph{conditional} decoder, our model can generate novel exemplars from seen/unseen classes, given their respective class attributes. These exemplars can subsequently be used to train any off-the-shelf classification model. One of the key aspects of our encoder-decoder architecture is a feedback-driven mechanism in which a discriminator (a multivariate regressor) learns to map the generated exemplars to the corresponding class attribute vectors, leading to an improved generator. Our model's ability to generate and leverage examples from unseen classes to train the classification model naturally helps to mitigate the bias towards predicting seen classes in generalized zero-shot learning settings. Through a comprehensive set of experiments, we show that our model outperforms several state-of-the-art methods, on several benchmark datasets, for both standard as well as generalized zero-shot learning.
\end{abstract}
\vspace{-2.5em}
\section{Introduction}
\footnotetext[1]{Equal contributions from both authors.}
The ability to correctly categorize objects from previously unseen classes is a key requirement in any truly autonomous object discovery system. Zero-shot Learning (ZSL) is a learning paradigm~\cite{DAP,akata2013label,xian2017zero} that tries to fulfil this desideratum by leveraging \emph{auxiliary} information that may be available for each seen/unseen class. ZSL models usually assume that this information is given in form of class attribute vectors or textual descriptions of classes. 

Typical approaches taken by existing ZSL models can be roughly categorized into the following: (1) Learning a mapping from the instance space to the class-attribute space and predicting the class of an unseen class test instance by finding its closest class-attribute vector~\cite{DAP,akata2013label}; (2) Defining the classifier for each unseen class as a weighted combination of the classifiers for the seen classes, where the combination weights are typically defined using similarity scores of unseen and seen class~\cite{norouzi2013zero,changpinyo2016synthesized}, and (3) Learning a probability distribution for each seen class and extrapolating to unseen class distributions using the class-attribute information~\cite{verma2017simple,guo2017synthesizing,wang2017zero}. A more detailed discussion of the related work on ZSL is provided in the Related Work section.
Although the aforementioned ZSL models have shown considerable promise on various benchmark datasets, a key limitation of most of these models is that, at test time, these are highly biased towards predicting the seen classes~\cite{chao2016empirical}. This is because the ZSL model is learned using labeled data only from the seen classes. Due to this issue, the ZSL models are usually evaluated in a restricted setting where the training and test classes are assumed to be disjoint, i.e., the test examples only come from the unseen classes and the search space is limited to the unseen classes only. The more challenging setting where the training and test classes are not disjoint is known as generalized zero-shot learning (GZSL), and is considered a more formidable problem setting. Recent work~\cite{chao2016empirical,xian2017zero} has shown that the accuracies of most ZSL approaches drops significantly in this setting.

In this work, we take a generative approach to the ZSL problem, which naturally helps address the generalized ZSL problem. Our approach is based on a generative model to \emph{synthesize} exemplars from the unseen classes (and, optionally, also from the seen classes), and subsequently training an off-the-shelf classification model using these synthesized exemplars. Our approach is motivated by, and is similar in spirit to, recent work on synthesizing exemplars for ZSL~\cite{BucherZSL,guo2017synthesizing,long2017zero,8066319}, which has shown to lead to improved performance, especially in the GZSL setting.
\begin{figure*}
\centering
\includegraphics[height=7cm,width=16cm]{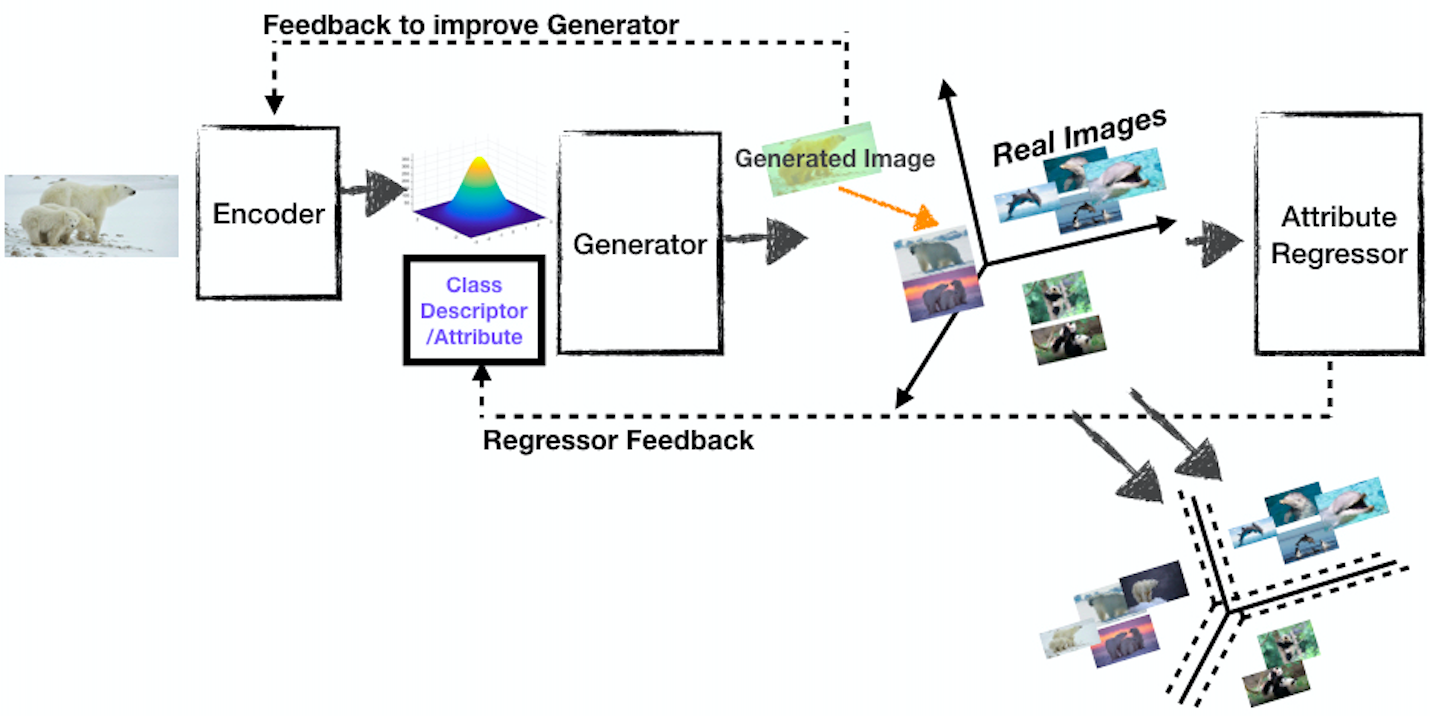}
\caption{An illustration of our model: Like a conditional VAE, the probabilistic encoder-generator reconstructs a noisy version of the original input(Generated Image). We add a feedback mechanism via loss from attribute regressor as well as the encoder, to improve the reconstruction capability of the decoder network. The dotted lines denote the feedback loss, while the orange line points to the desired location of the reconstruction in feature space. At test time, predictions are made using an off-the-shelf classifier (e.g. SVM) trained on synthesized examples. The 3D coordinate is representative of feature space and right bottom parts represents max-margin classifier}
\label{fig:pictorial}
\vspace{-1em}
\end{figure*}

For exemplar generation, we develop a generative model based on a \emph{conditional} variational autoencoder (VAE) architecture, in which the latent code of any instance is augmented with the class-attribute vector. This architecture is further coupled with a \emph{discriminator} (a multivariate regressor) that learns a mapping from the VAE generator's output to the class-attribute. This feedback helps to improve the generator by encouraging it to generate exemplars that are of highly discriminative nature. Moreover, the discriminator also allows us to operate in semi-supervised settings by incorporating unlabeled examples for which we do not know the class label (and thus no class-attributes are available for these examples). Once the model has been learned, it can be used to generate exemplars for any unseen class (and, optionally, also seen classes), given its class-attributes, and the exemplars can be used as labeled training examples to train any off-the-shelf classification model. Notably, since the classification model is trained using labeled examples from seen as well as unseen classes, at test time, it is not biased towards predicting seen classes in the GZSL setting, as also evidenced by recent work on leveraging synthesized unseen class examples~\cite{BucherZSL,guo2017synthesizing,long2017zero,8066319}. The model in \cite{BucherZSL} is a vanilla conditional generative
model, while ours is based on an explicit feedback driven
mechanism. This leads to a very different model architecture and inference,
and yields much better prediction accuracies. \cite{guo2017synthesizing} represented
each unseen class by Gaussian distribution while
\cite{long2017zero,8066319} based on semantic-visual embedding with Diffusion Regularization.

Also note that, in our proposed framework, the final classification stage directly predicts the actual \emph{class label} for each test example, as opposed to predicting the \emph{class-attribute vector}~\cite{socher2013zero,akata2013label}, which further necessitates a nearest neighbor search to find the class label. This is appealing because the nearest neighbor search approach, as is commonly used in most ZSL methods, is known to suffers from issues, such as the hubness problem~\cite{hubness}.


\section{Background and Notation}\label{sec:methodology}

In the ZSL problem, we assume that there are $S$ seen classes for which we have labeled training examples and $U$ unseen classes for which we do not have any labeled training examples. The test examples can either be exclusively from unseen classes (the traditional ZSL setting) or from both seen and unseen classes (the generalized ZSL setting~\cite{chao2016empirical,xian2017zero}). In this work, our focus will be on the GZSL setting, although our model applies to both settings (we provide comprehensive experimental evaluations for both).

Although we do not have access to any labeled examples from the unseen classes, for each (seen/unseen) class $c=1,\ldots,S+U$, we assume that we are given the respective class-attribute vectors $\{\mathbf{a}_c\}_{c=1}^{S+U}$, where the attribute vector of class $c$ is $\mathbf{a}_c \in \mathbb{R}^L$. ZSL models are usually based on leveraging the class-attribute information to transfer information from seen classes to the unseen classes.

Note that each seen class labeled training example $\{\xv_n,y_n\}$ can be equivalently denoted as $\{\xv_n,\av_{y_n}\}$, the feature vector and class-attribute vector pair for this example. Therefore, assuming a total of $N_S$ examples from the seen classes, the training data from seen classes can be collectively denoted by $\Dcal_S = \{\xv_n,\av_{y_n}\}_{n=1}^{N_S}$. The goal in ZSL is to learn a classification model using $\Dcal_S$ and then use the learned model to predict the labels for test examples.

\section{The Basic Model}

Our model, shown via the pictorial illustration in Fig.~\ref{fig:pictorial} is based on a variational autoencoder(VAE) architecture~\cite{kingma2014vae}. The VAE consists of a probabilistic encoder model with parameters $\theta_E$ and a probabilistic decoder (a.k.a. generator) model with parameters $\theta_G$. In our model, the generator is also conditioned on the class-attribute vector, which enables us to synthesize exemplars from any class $c$ by simply specifying the corresponding class-attribute vector $\av_c$, along with an unstructured code $\zv$. 


Note that this architecture is akin to a conditional VAE model~\cite{sohn2015learning}, except for some key differences
\begin{itemize}
	\item We assume that the latent code and the class-attribute are \emph{disentangled} (latent code $\zv_n$ represents the unstructured part of $\xv_n$ and the class-attribute vector $\av_{y_n}$ represents the class-specific discriminative information); this helps in generating exemplars that are high discriminative in nature, as guided by the class-attribute vector.
    \item The model also consists of a mapping from the decoder's output to the class-attribute vector; this mapping is learned via a discriminator which is a multivariate regression model with parameters $\theta_R$ that maps the decoder's output $\hat{\xv}_n$ to the respective class-attribute vector $\av_{y_n}$ via a feedforward network (learned jointly with the rest of the model). The regressor plays two key roles in our model: (1) It provides feedback to the generator (more on this in Sec.~\ref{sec:disc}), which results in generation of exemplars that can be discriminated easily; and (2) It allows using unlabeled examples during training by computing the probability distribution $p(\av|\xv)$ on their class-attribute vector.
\end{itemize}

Note that, in our proposed model, each example $\xv_n$ is influenced by two sources - the latent code $\zv_n$ which represents the unstructured (class-independent) component, and the class-attribute vector $\av_{y_n}$ which represents the structured (class-specific) component.  We describe each of the model components in more detail in Section~\ref{sec:modelarch}.


\subsection{Training the Final Classifier} 

Once the generative model has been learned, we can generate \emph{labeled} exemplars of any class by first generating the unstructured component $\zv$ randomly from the prior $p(\zv)$, specifying the class $c$ (via the class attribute vector $\av_c$) of the exemplar to be generated, and then generating the example $\xv$ using the generator. We generate a fixed number of exemplars from each class and these generated exemplars are finally used to train a discriminative classifier, such as a support vector machine (SVM) or a softmax classifier. Since this stage utilizes labeled examples from both seen and unseen classes, our approach is inherently robust against the bias towards seen classes, as also evidenced by other recent work~\cite{BucherZSL,guo2017synthesizing}. Also note that, for training this classifier, we can either use the original labeled examples from the seen classes or can also augmented those examples using additional exemplars from the seen classes as well.  

\section{The Complete Model Architecture}
\label{sec:modelarch}

\begin{figure}[!htbp]
    \centering
    \includegraphics[width=0.45\textwidth]{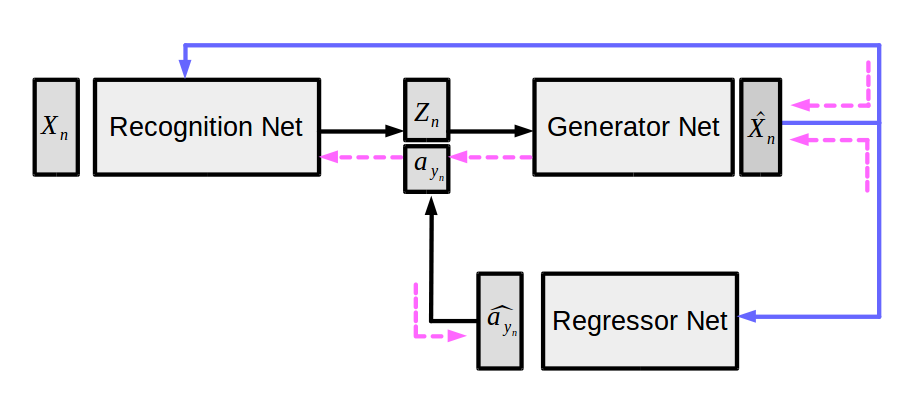}
    \caption{The proposed architecture for zero-shot set-up. Each block represents a feed-forward neural network. The encoder to $\mathbf{z}_n$ link is stochastic similar to a VAE. The blue lines direct feedback connection into regressor and recognition network for the generated $\hat{X}_n$. The red-lines represent the back propagation direction.} 
    \label{fig:archi}
    \vspace{-1em}
\end{figure}

We now describe our model architecture in more detail. Our model, as shown through the block-diagram in Fig.~\ref{fig:archi}, is based on a variational autoencoder (VAE), consisting of an encoder $p(\zv|\xv)$ (a recognition network) and a decoder/generator $p(\xv|\zv,\av)$. Note that, unlike the standard VAE, the generator's input consists of both the latent code $\zv$ as well as the class-attribute vector $\av$, similar to a conditional VAE (CVAE) architecture~\cite{sohn2015learning}, enabling the generated exemplars from different classes to be distinguishable from each other. However, as compared to the traditional CVAE, our architecture has a few key differences, motivated by the goal is having a generator that can generate exemplars from any given class that can act as a surrogate to the real examples that class:

\begin{itemize}
	\item It consists of a discriminator (a multi-output regressor, which we call regressor net) that learns to map a real example $\xv$ (from seen classes) or a generator-synthesized example $\hat{\xv}$ (from unseen/seen classes) to the corresponding class-attribute vector $\av$. Backpropagating the regressor's loss helps to improve the generator by ensuring that the generated exemplars are representative of the associated class.
    \item The regressor also enables using our model in a semi-supervised setting, where some training examples do not have labels/class-attribute vectors. For such examples, the class-attribute vector is replaced by the output distribution $p(\av|\xv)$ of the regressor (Sec.~\ref{sec:disc}).
    \item A link from the generator back to the recognition model (encoder) to ensure that the generator's output $\hat{x}$ is as good as the actual input $\xv$, i.e., the distribution $p(\zv|\hat{x})$ is closed to the distribution $p(\zv|\xv)$.
\end{itemize}
\vspace{-1em}



Our model architecture draws its inspiration from recent work on controllable text generation~\cite{hu2017controllable}, where the goal is to generate text having a certain desired characteristics, such as positive/negative sentiment, by specifying a binary attribute. In contrast, in our ZSL setting, the VAE model is conditioned on the class-attribute vector, enabling us to smoothly transition the generation from seen to unseen classes by varying the class-attribute vector. Moreover, while the focus of the work in~\cite{hu2017controllable} was on text generation, our goal here is to leverage such a  framework to solve the generalized ZSL problem.

Next, we briefly describe the key components of our model architecture shown in Fig.~\ref{fig:archi}.
\subsection{The Discriminator/Regressor}
\label{sec:disc}
Our discriminator/regressor, defined by a probabilistic model $p_R(\av|\xv)$ with parameters $\theta_R$, is a feedforward neural network that learns to map the example $\xv \in \mathbb{R}^D$ to its corresponding class-attribute vector $\av \in \mathbb{R}^L$. The regressor is learned using two sources of data: 

\begin{itemize} 
	\item Labeled examples $\{\xv_n,\av_{y_n}\}_{n=1}^{N_S}$ from the seen class, on which we can define a supervised loss, given by
\begin{equation}
\mathcal{L}_{Sup} (\theta_R) = -\mathbb{E}_{\{\xv_n,\av_{y_n}\}}[p_R(\av_{y_n}|\xv_n)]
\end{equation}
	\item Synthesized examples $\hat{\xv}$ from the generator, for which we can define an unsupervised loss, given by 
\begin{equation}
    \mathcal{L}_{Unsup}(\theta_R) = -\mathbb{E}_{p_{\theta_G}(\mathbf{\hat{x}}|\mathbf{z},\mathbf{a})p(\mathbf{z})p(\mathbf{a})}[p_R(\mathbf{a}|\mathbf{\hat{x}})]
\end{equation}
Note that the unsupervised loss is computed by taking a latent code $\zv$ sampled from the prior $p(\zv)$, along with a class-attribute vector $\av$ sampled from the empirical distribution $p(\av)$, generating an exemplar from the generator distribution $p_{\theta_G}(\mathbf{\hat{x}}|\mathbf{z},\mathbf{a})$, and then taking an expectation w.r.t. these distributions.  During this phase, we can use attributes both from seen as well as unseen classes. 

The overall training objective of the regressor is then defined as the following weighted combination of the supervised and the unsupervised training objectives
\begin{equation}
    \min_{\theta_{R}} \mathcal{L}_R = \mathcal{L}_{Sup} + \lambda_R \cdot \mathcal{L}_{Unsup}
    \label{eq:disc_total}
\end{equation}
The above optimization problem is the first step of the alternating optimization procedure. It optimizes the regressor parameters $\theta_R$ to make the regressor predict the correct class-attribute vector even with a noisy signal $\hat{\mathbf{x}}$. Note that, in this step, we assume that the generator distribution is fixed. 
\end{itemize}




\subsection{The Encoder and Conditional Generator}
The conditional VAE in our architecture (Fig.~\ref{fig:archi}) consists of the conditional generator $p_G(\xv|\zv,\av)$ with parameters $\theta_G$, which are responsible for generating the exemplars that will subsequently be used to train the final classification model. Hence the training needs to be designed such that the generated class conditional distribution nicely approximates the true distribution. Denoting the VAE encoder as $p_E(\zv|\xv)$ with parameters $\theta_E$,  and the  regressor output distribution as $p_R(\av|\xv)$, the VAE loss function is given by (assuming the regressor to be fixed)
\begin{equation}
\begin{aligned}
	\mathcal{L}_{VAE}(\theta_E,\theta_G) &= -\mathbb{E}_{p_{E}(\mathbf{z}|\mathbf{x}),p(\mathbf{a}|\mathbf{x})} [\log p_{{G}}(\mathbf{x}|\mathbf{z},\mathbf{a})]\\ &+ \text{KL}(p_{E}(\mathbf{z}|\mathbf{x})||p(\mathbf{z}))
\end{aligned}
\label{eq:VAE}
\end{equation}
where the first term on the R.H.S. is the generator's reconstruction error and the second term promotes the VAE posterior (the encoder) to be close to the prior.

We model the VAE encoder $p_E(\zv|\xv)$, VAE conditional decoder/generator $p_G(\xv|\zv,\av)$, and the regressor $p_R(\av|\xv)$ as Gaussian distributions. Also note that the factorization of the joint distribution over overall latent code $(\zv,\av)$ into two components $p_{E}(\mathbf{z}|\mathbf{x})$ and $p_{R}(\mathbf{a}|\mathbf{x})$ is consistent with our attempt at learning a disentangled representation~\cite{hu2017controllable}. 

\textbf{Discriminator-Driven Learning:} As described in Sec.~\ref{sec:disc}, we use the discriminator/regressor to improve the generator by backpropagating its error that encourages generation of exemplars, $\mathbf{\hat{x}}$ coherent with the class-attribute $\mathbf{a}$. We perform this by using a couple of loss functions. The first one simply assumes that the regressor, has optimal parameters and any reason for it not regressing to the correct value is because of the poor generation by the generator:
\begin{equation}
    \mathcal{L}_{c}(\theta_{G}) = -\mathbb{E}_{p_G(\mathbf{\hat{x}}|\mathbf{z},\mathbf{a})p(\mathbf{z})p(\mathbf{a})}[\log p_{R}(\mathbf{a}|\hat{\xv})]
\end{equation}
This loss encourages the generator to create samples such that the regressed attribute vector by the discriminator is correct. We also add an additional term that acts as a regularizer that encourages the generator to generate a good class-specific sample even from a random $\mathbf{z}$ drawn from the prior $p(\zv)$ and combined with class-attribute from $p(\av)$. This is akin to doing semi-supervised learning.
\begin{equation}
    \mathcal{L}_{Reg}(\theta_G) = -\mathbb{E}_{p(\mathbf{z})p(\mathbf{a})}[\log p_G(\mathbf{\hat{x}}|\mathbf{z},\mathbf{a})]
\end{equation}
The above ensures that the quality of the synthesized exemplars, to be used to train the final classifier, is at par with that of the true data. While the above two loss functions help us increase the coherence of $\mathbf{\hat{x}} \sim p_G(\mathbf{\hat{x}}|\mathbf{z},\mathbf{a})$ with the class-attribute $\mathbf{a}$, we also need to enforce the independence (disentanglement)~\cite{hu2017controllable} of the unstructured component $\mathbf{z}$ from the class-attribute $\mathbf{a}$. To this end, we use the encoder to ensure that the sampling distribution and the one obtained from the generated exemplar follow the same distribution. More formally, we add a loss component,
\begin{equation}
    \begin{aligned}
    \mathcal{L}_{E}(\theta_G) = -\mathbb{E}_{\mathbf{\hat{x}}\sim p_G(\mathbf{\hat{x}}|\mathbf{z},\mathbf{a})}\text{KL}[(p_{E}(\mathbf{z}|\mathbf{\hat{x}})||q(\mathbf{z}))]
    \end{aligned}
\end{equation}
The distribution $q(\mathbf{z})$ could be the prior $p(\mathbf{z})$ or the posterior from a labeled sample $p(\mathbf{z}|\mathbf{x}_n)$, in which case the attribute component $\mathbf{a}_{y_n}$ is used. Hence the complete learning objective for the generator and encoder is given by,
\begin{equation}
    \begin{aligned}
    \min_{\theta_G,\theta_E}  \mathcal{L}_{VAE} + \lambda_c \cdot \mathcal{L}_{c} + \lambda_{reg} \cdot \mathcal{L}_{Reg} + \lambda_E \cdot \mathcal{L}_E \\
    \end{aligned}
    \label{eq:EG_total}
\end{equation}
The overall learning objective for the conditional auto-encoder is a weighted combination of all the components discussed above, effectively optimizing parameters to ensure good density estimation. Vanilla conditional generative models like conditional
VAE \cite{mishra2017generative} and \cite{BucherZSL} crucially lack such a mechanism. The weight are hyperparameters, tuned for optimal reconstruction. This completes the second step of the alternating optimization. 
\section{Related Work}
\label{sec:relwork}

Zero-shot learning (ZSL) has received a significant amount of interest recently. Due to lack of space, it will not be possible to comprehensively cover all the work in this area. In this section, we provide an overview of some of the representative methods, both for traditional ZSL as well as generalized ZSL.

Some of the early works on ZSL were based on learning a direct/indirect mapping~\cite{DAP} from the instances to the class-attributes. This mapping is then applied on the test data to first predict the class-attribute vector, and used to predict the class by finding the most similar attribute vector.

Another popular approach for ZSL is based on learning a shared embedding of seen and unseen class instances into the class-attribute space~\cite{socher2013zero,norouzi2013zero}. After projection, nearest neighbor methods can be used to find the most similar class attribute vector for the (projected) test instance, which corresponds to the most likely class. While conceptually simple and easy to implement, these methods suffer from shortcomings such as the hubness problem~\cite{radovanovic2010hubs}. 

In a similar vein of embedding instances to an attribute/semantic space, some other approaches learn a mapping of instances to a semantic (attribute) space, with \cite{akata2013label} learning a bilinear compatibility function between the instances and attribute space using ranking loss and \cite{frome2013devise} optimizing the structural SVM loss to learn the bilinear compatibility. Embedding based methods for ZSL have also been extended to learn  non-linear  multi-modal embeddings. 


Using the fact that the class-attributes can be used to compute relationships between seen and unseen classes (e.g., using a class-attribute based similarity measures), a number of ZSL methods have been proposed that are based on representing the parameters representing each unseen class as a similarity-weighted combination of the parameters representing the seen classes~\cite{norouzi2013zero,romera2015embarrassingly,changpinyo2016synthesized}.

The generalized ZSL problem~\cite{chao2016empirical,xian2017zero} where the training and test classes are not disjoint is considerably more challenging as compared to the traditional ZSL, and a recent focus has been to design ZSL methods that can work robustly in this setting without being biased towards predicting seen classes. Generative models~\cite{verma2017simple,guo2017synthesizing,BucherZSL,wang2017zero} are promising in this setting. One of the ways these models can solve the GZSL problem is by generating synthetic labeled examples from the unseen classes and then using these examples (and the labeled examples from other seen classes) to train a classification model.

Following this approach, and in a similar spirit to our work, a number of recent works~\cite{guo2017synthesizing,BucherZSL} have tried to use synthesized examples both for the seen as well as unseen classes to perform the generalized zero-shot task. \cite{guo2017synthesizing} synthesize samples for each class by approximating the class conditional distribution of the unseen classes based on the learned class conditional distributions of the seen classes and their corresponding attribute vectors. On the other hand \cite{BucherZSL} perform adversarial training to train generators and use the domain adapted samples for perform classification.

Finally, the ability to generate exemplars from unseen class and use them in training classification models can also help mitigate the domain-shift problem~\cite{kodirov2015unsupervisedDA} encountered by traditional ZSL methods if the distribution of seen classes and unseen classes are not the same. Given labeled examples from the seen classes and the synthetic labeled examples from the unseen classes, supervised/semi-supervised domain adaptation methods can be readily applied to address the domain shift problem. 

Despite the significant amount of progress in ZSL over the past few years, we would also like to point out that differences in evaluation protocols for evaluating ZSL models often make it hard to have a fair comparison between the various methods. In a recent work,~\cite{xian2017zero} lay down a set of guidelines on choosing data-splits and evaluations protocols to ensure fair comparisons. Our experimental settings strictly adhere to these guidelines as much as possible.

\section{Experiments}
\label{sec:expt}

To test the effectiveness of our model (referred to as \textbf{SE-GZSL} for Synthesized Examples for Generalized Zero-Shot Learning), we conduct an extensive evaluation on several benchmark datasets and compare it with various state-of-the-art ZSL models. Note that the baselines also include some recently proposed methods based on exemplar generation~\cite{BucherZSL,mishra2017generative,guo2017synthesizing}. We report our results on the following benchmark datasets, while also following the guidelines offered by~\cite{xian2017zero} for evaluating ZSL models:

\begin{itemize}
 \item \textbf{Animals with Attributes}: The AwA \cite{lampert2009learning} dataset  contains 30,475 images with a standard split of 40 seen classes (training set) and 10 unseen classes (test set). Each class has a human-provided 85-dimensional class-attribute vector. Since raw images for the original dataset were not available, we used the VGG19 features. Recently, an updated version of the dataset with raw images has been made available. For completeness, we evaluate our model on both the datasets, henceforth referred to as AwA1 \cite{lampert2009learning} and AwA2 \cite{xian2017zero}.
 \item \textbf{SUN Scene Recognition}: The SUN \cite{xiao2010sun} dataset comprises of 717 scenes. For the ZSL setting, we use the widely used split of 645 seen classes with 72 unseen classes. This dataset has 14,340 fine-grained images, with attributes available at the image level. We combine the attributes of all images in a class to obtain class-level attributes and use them for our training.
 \item \textbf{Caltech UCSD Birds 200} : The CUB dataset  \cite{welinder2010caltech} consists of 200 classes with 11,788 fine-grained images of birds. We use the given split of 150 unseen and 50 seen classes. Like the SUN dataset, this one too has attributes available at image level and we average them across each class to get class-attributes. 
 \item \textbf{Imagenet}: We also evaluate the zero-shot classification accuracy on the large-scale Imagenet dataset. The setup here involves training using the images from the 1000 class ILSVRC 2012 \cite{DBLP:journals/corr/RussakovskyDSKSMHKKBBF14} data and testing on the non-overlapping 360 classes from the ILSVRC 2010 data. Unlike other datasets, we use the GoogLeNet \cite{googlenet} features for this dataset.
\end{itemize}

Datasets and their statistics are summarized in Table~\ref{tab:data}. While human-curated attributes for each class are available for most datasets, we use word2vec representations of each class as the class-attribute vector for Imagenet.
\begin{table}[h!]
\small
\centering
 \addtolength{\tabcolsep}{-2.0pt}
 \begin{tabular}{||c|c|c|c||} 
 \hline
 Dataset & Attribute/Dim & \#Image & Seen/Unseen Class \\ [0.5ex] 
 \hline\hline
AWA1 & A/85 & 30475 & 40/10 \\
AWA2 & A/85 & 37322 & 40/10 \\
CUB & A/312 & 11788 & 150/50 \\
SUN & A/102 & 14340 & 645/72 \\
Imagenet & W/1000 & 254000 & 1000/360 \\

 \hline

 \hline
 \end{tabular}
 \vskip 10pt
 \caption{Datasets used in our experiments, and their statistics} \label{tab:data}
 \vspace{-1em}
\end{table}

\subsection{Parameter Settings and Evaluation}
For each of the mentioned datasets, we evaluate our method on the commonly used standard split as well as on the split proposed in recent work~\cite{xian2017zero} about evaluation protocols for ZSL. A new version of the AwA dataset was also proposed in ~\cite{xian2017zero}. We therefore report our results on both the old AwA1 dataset,  as well as the new AwA2 dataset. For each of the datasets we use the ResNet features of the images. No extra fine-tuning was done to improve the image features. The network was optimized based on the loss function discussed in Sec.~\ref{sec:modelarch}, using the Adam \cite{kingma2014adam} optimizer. The learning started with pre-training the VAE using the loss from Eq.~\ref{eq:VAE}. This is followed by joint alternating-training of regressor and encoder-generator pair by optimizing the loss from Eq.~\ref{eq:disc_total} and Eq.~\ref{eq:EG_total}, until convergence. The hyerparameters were chosen based on a train-validation split and were used while training the model on complete data. Though there may be some small variability in the results based on the values for the hyperparameters, we used $\lambda_{R} = 0.1$, $\lambda_{c} = 0.1$, $\lambda_{reg} = 0.1$  and $\lambda_{E} = 0.1$, which worked well for most of the experiments. It is important to note that the same architecture is used across all datasets and no extra data/feature engineering is used/performed to improve accuracies, thus showcasing the efficacy of the training schedule for this task. The encoder is realised using a two-hidden layer feedforward network while the decoder and the regressor are modeled as feedforward networks consisting of one hidden layer with 512 hidden units each. 

For the evaluation criteria, we use the average per-class accuracy.
This metric reduces the bias of classes having higher examples in the test set, and provides a better measure of the model performance~\cite{xian2017zero}.

We shall now discuss the experimental setup for the two settings we experiment with: ZSL and GZSL. Let the datasets be available in two parts, the labeled examples from the seen classes, $\mathcal{X}^S$ and the unlabeled examples from the unseen classes, $\mathcal{X}^U$.

\begin{table*}[!htbp]
\small
 \centering
 \addtolength{\tabcolsep}{-3pt}
 \begin{tabular}{|c|c c c | c c c | c c c | c c c |} 
 \hline
  & \multicolumn{3}{c|}{\textbf{SUN}} & \multicolumn{3}{c|}{\textbf{CUB}} & \multicolumn{3}{c|}{\textbf{AWA1}} & \multicolumn{3}{c|}{\textbf{AWA2}} \\ \hline
 \textbf{Method} &\textbf{U $\rightarrow$ S+U}& \textbf{S $\rightarrow$ S+U} & \textbf{H} & \textbf{U $\rightarrow$ S+U}& \textbf{S $\rightarrow$ S+U} & \textbf{H} & \textbf{U $\rightarrow$ S+U}& \textbf{S $\rightarrow$ S+U} & \textbf{H} & \textbf{U $\rightarrow$ S+U}& \textbf{S $\rightarrow$ S+U} & \textbf{H} \\ 
  \hline
   
  \textbf{CONSE} \cite{norouzi2013zero} & 6.8 & 39.9 & 11.6 & 1.6& \textbf{72.2}& 3.1& 0.4& \textbf{88.6}& 0.8& 0.5& \textbf{90.6}& 1.0 \\
  \textbf{CMT*} \cite{socher2013zero} & 8.1& 21.8& 11.8& 7.2& 49.8& 12.6& 0.9& 87.6& 1.8& 0.5& 90.0& 1.0 \\
  \textbf{SSE} \cite{saligram2016learningJoint} & 2.1& 36.4& 4.0& 8.5& 46.9& 14.4& 7.0& 80.5& 12.9& 8.1& 82.5& 14.8 \\
  \textbf{SJE} \cite{SJE} & 14.7& 30.5& 19.8& 23.5& 59.2& 33.6& 11.3& 74.6& 19.6& 8.0& 73.9& 14.4\\
  \textbf{ESZSL} \cite{romera2015embarrassingly} & 11.0& 27.9& 15.8& 12.6& 63.8& 21.0& 6.6& 75.6& 12.1& 5.9& 77.8& 11.0 \\
  \textbf{SYNC}\cite{changpinyo2016synthesized}& 7.9& \textbf{43.3}& 13.4& 11.5& 70.9& 19.8& 8.9& 87.3& 16.2& 10.0& 90.5& 18.0 \\
  \textbf{SAE} \cite{SAE2017} & 8.8& 18.0& 11.8& 7.8& 54.0& 13.6& 1.8& 77.1& 3.5& 1.1& 82.2& 2.2 \\
  \textbf{LATEM} \cite{xian2016latent}& 14.7& 28.8& 19.5& 15.2& 57.3& 24.0& 7.3& 71.7& 13.3& 11.5& 77.3& 20.0  \\
  \textbf{ALE} \cite{akata2013label} & 21.8&  33.1&  26.3&  23.7&  62.8&  34.4&  16.8&  76.1&  27.5&  14.0&  81.8&  23.9 \\
  \textbf{DEVISE} \cite{frome2013devise} & 16.9& 27.4& 20.9& 23.8& 53.0& 32.8& 13.4& 68.7& 22.4& 17.1& 74.7& 27.8 \\
  \textbf{CVAE-ZSL}\cite{mishra2017generative}&-- & --& 26.7 &--&-- & 34.5 &--&-- & 47.2 & --&-- & 51.2 \\
 \hline
 \textbf{SE-GZSL} (Ours) & \textbf{40.9} & {30.5} & \textbf{34.9} & \textbf{41.5} & {53.3} & \textbf{46.7} & \textbf{56.3} & {67.8}& \textbf{61.5} & \textbf{58.3} & {68.1}  &  \textbf{62.8}\\
 \hline
 \end{tabular}
 \vskip 8pt
 \caption{Accuracy for GZSL, on proposed split(PS). U and S represents top-1 accuracy on unseen and seen class. H: Harmonic mean.}
 \label{unsgzsl}
 \vspace{-1em}
\end{table*}

\subsection{Generalized Zero Shot Learning}
\label{sec:gzsl}
The GZSL setting involves performing classification when the test set has examples from both the seen and the unseen classes, with no prior distinction between them. For this, we perform an 80-20 random split of the dataset to obtain, $\mathcal{X}_{train}^S$ and $\mathcal{X}_{test}^S$. The split is done ensuring that there are some examples from each of the $S$ classes. 
We train our model on $X_{train}^S$. Once trained, samples are synthesized for all the $S+U$ classes using our generative model. These samples finally used to train a multi-class linear SVM. The examples from $\mathcal{X}^U$ (referred as $\mathcal{Y}^{tr}$) and $\mathcal{X}_{test}^S$ (referred as $\mathcal{Y}^{ts}$) are then used to calculate the average per-class accuracy. For the GZSL setting, the evaluation measures are denoted as
\begin{itemize}
\item $Acc_{\mathcal{Y}^{tr}}$ : $S \rightarrow S+U$ : Average per-class classification accuracy on $\mathcal{X}_{test}^S$ using a classifier trained for $S+U$
\item $Acc_{\mathcal{Y}^{ts}}$ : $U \rightarrow S+U$ : Average per-class classification accuracy on $\mathcal{X}_{test}^U$ using a classifier trained for $S+U$
\end{itemize}
To mitigate the bias towards seen classes accuracy, we evaluate the harmonic mean of the above defined average per-class top-1 accuracies as 
$
\mathbf{H} = (2\cdot Acc_{\mathcal{Y}^{tr}} \cdot Acc_{\mathcal{Y}^{ts}})/({Acc_{\mathcal{Y}^{tr}} + Acc_{\mathcal{Y}^{ts}}})
$

The results for the test accuracy on the unseen classes and the aggregate measure (harmonic mean), for this setup are compiled in Tables~\ref{unsgzsl}. The number of samples synthesized is a hyper-parameter and can be chosen to balance evaluation time and accuracy. The results clearly demonstrate that our model can significantly mitigate the GZSL issue of the bias towards seen classes, which a number of previous ZSL approaches tend to suffer~\cite{chao2016empirical,xian2017zero}. Our approach outperforms previous approaches on both  the unseen class test accuracy, $Acc_{\mathcal{Y}^{ts}}$ as well as the harmonic mean measure $\mathbf{H}$, which quantities the aggregate performance across both seen and unseen test classes. Here for the SVM training we used the different weight for the seen and unseen class. Using the validation data we found that linear SVM with the C=1 outperforms w.r.t. other hyper-parameter. Here seen weight $1.0$ while unseen weight is $0.2,0.2,0.05$ and $0.05$ is used for the SUN, CUB, AWA1 and AWA2 respectively.

\subsection{Conventional Zero Shot Learning}
\label{sec:zsl}
For the conventional ZSL setting, we first train our generative model on $\mathcal{X}^S$. We then synthesize samples for the unseen classes and finally train a multi-class linear SVM using the generated training data from these unseen classes. The SVM is used to predict the classes for the test examples $\mathcal{X}^U$. The average per-class accuracy, $Acc_{\mathcal{Y}^{ts}}$ is reported in Table \ref{tab:zsl} and Table \ref{imgnt}. The improvements are consistent across scale and the complexity of images in the dataset. This is evident from the improvement on the large-scale Imagenet dataset as well complex fine-grained datasets like CUB. The large number of classes and relatively fewer training examples per class in the SUN dataset do not hamper the performance of our method. 

To further probe the efficacy of the feedback mechanism, we perform an ablation study, where the model is trained without feedback mechanism for the conventional ZSL setting. The results in Table.~\ref{tab:zsl}  cleary demonstrate the benefits of the feedback mechanism and the carefully designed loss function. This also explicates why our model outperforms other recently proposed architectures, such as \cite{BucherZSL,mishra2017generative} which don’t have a feedback-driven mechanism in their generative models. The degrade in performance without feedback is particularly significant in fine-grained datasets like CUB.

\begin{table*}[!htbp]
\small
 \centering
 \addtolength{\tabcolsep}{8.0pt}
 \begin{tabular}{|c| c c | c c | c c | c c |} 
 \hline
  & \multicolumn{2}{c|}{\textbf{SUN}} & \multicolumn{2}{c|}{\textbf{CUB}} & \multicolumn{2}{c|}{\textbf{AWA1}} & \multicolumn{2}{c|}{\textbf{AWA2}} \\ \hline
 \textbf{Method} & \textbf{SS} & \textbf{PS} & \textbf{SS} & \textbf{PS} & \textbf{SS} & \textbf{PS} & \textbf{SS} & \textbf{PS} \\ 
  \hline
  \textbf{CONSE} \cite{norouzi2013zero} & 44.2 & 38.8 & 36.7 & 34.3 & 63.6 & 45.6 & 67.9 & 44.5 \\
  \textbf{SSE} \cite{saligram2016learningJoint} & 54.5 & 51.5 & 43.7 & 43.9 & 68.8 & 60.1 & 67.5 & 61\\
  \textbf{LATEM} \cite{xian2016latent} & 56.9 & 55.3 & 49.4 & 49.3 & 74.8 & 55.1 & 68.7 & 55.8\\
 \textbf{ALE} \cite{akata2013label} & 59.1 & 58.1 & 53.2 & 54.9 & 78.6 & 59.9 & 80.3 & 62.5 \\
 \textbf{DEVISE} \cite{frome2013devise} & 57.5 & 56.5 & 53.2 & 52.0 & 72.9 & 54.2 & 68.6 & 59.7 \\
 \textbf{SJE} \cite{SJE} & 57.1 & 53.7 & 55.3 & 53.9 & 76.7 & 65.6 & 69.5 & 61.9 \\
 \textbf{ESZSL} \cite{romera2015embarrassingly} & 57.3 & 54.5 & 55.1 & 53.9 & 74.7 & 58.2 & 75.6 & 58.6 \\
 \textbf{SYNC}\cite{changpinyo2016synthesized} & 59.1 & 56.3 & 54.1 & 55.6 & 72.2 & 54.0 & 71.2 & 46.6\\
 
 \textbf{SAE} \cite{SAE2017} & 42.4 & 40.3 & 33.4 & 33.3 & 80.6 & 53.0 & 80.2 & 54.1 \\
 
\textbf{SSZSL} \cite{guo2017synthesizing} & -- & -- & 55.75 & -- & 82.67 & -- & -- & -- \\
\textbf{GVRZSC} \cite{BucherZSL} & --  & -- & 60.1 & -- & 77.1 & -- & -- & -- \\
\textbf{GFZSL}\cite{verma2017simple} & 62.9 & 62.6 & 53.0  & 49.2 & 80.5 & 69.4 & 79.3 & 67.0 \\
 \textbf{CVAE-ZSL}\cite{mishra2017generative} & -- & 61.7 & -- & 52.1 & -- & \textbf{71.4} & -- & 65.8 \\
 \hline
 \textbf{SE-ZSL} (Without Feedback) & {62.0} & {61.2} & {59.8} & {54.1} & {78.4} & {68.2} & {79.3} & {66.3}\\
 \textbf{SE-ZSL} (Ours) & \textbf{64.5} & \textbf{63.4} & \textbf{60.3} & \textbf{59.6} & \textbf{83.8} & {69.5} & \textbf{80.8} &  \textbf{69.2}\\
 \hline
 \end{tabular}
 \vskip 8pt
 \caption{Zero Shot Learning Accuracy on the SUN, CUB, AWA1 and AWA2 dataset. Here SS stands for the Stranded Split for each dataset that has been in use previously and PS is the new proposed split by \cite{xian2017zero}. We also experimented on a 707/10 split for SUN \cite{SAE2017} and achieved accuracy $93.5\%$. Also in transductive setting GFZSL\cite{verma2017simple} has the $63.9\%, 51.2\%, 84.9\%$ and $ 81.0\%$ accuracy on the SUN, CUB, AWA1 and AwA2 datasets respectively in the PS setting.}
 \label{tab:zsl}
 \vspace{-1em}
\end{table*}

\begin{table}[h!]
\small
\centering
 \begin{tabular}{||c| c ||} 
 \hline
 Method & Accuracy \\ [0.5ex] 
 \hline\hline
  AMP\cite{fu2015zero} & 13.1 \\ 
 DeViSE:\cite{frome2013devise} & 12.8 \\
 ConSE \cite{norouzi2013zero} & 15.5 \\
 SS-Voc \cite{fu2016semi} & 16.8 \\
 {CVAE-ZSL}\cite{mishra2017generative} & 24.7 \\ [1ex] 
 \hline
 \textbf{SE-ZSL} (Ours) & \textbf{25.43}\\
 \hline
 \end{tabular}
 \vskip 10pt
 \caption{Per-class accuracy of ZSL for the ImageNet dataset. 1000 class ILSVRC-12 are used for training and ILSVRC-10(class that are not present in ILSVRC-12) are used for testing}\label{imgnt}
 \vspace{-1em}
\end{table}

\subsection{Quality of Synthesized Samples}
Our quantitative results reported for GZSL (Sec.~\ref{sec:gzsl}) and ZSL (Sec.~\ref{sec:zsl}) demonstrate that the samples generated by our model are of good quality and effective for classification tasks. To gain a further insight into the quality of the generated samples, we compare the empirical distribution of the generated samples from a few unseen classes to the empirical distribution of the real samples from
\begin{figure}[!htbp]
\centering
\includegraphics[height=4.5cm,width=6cm]{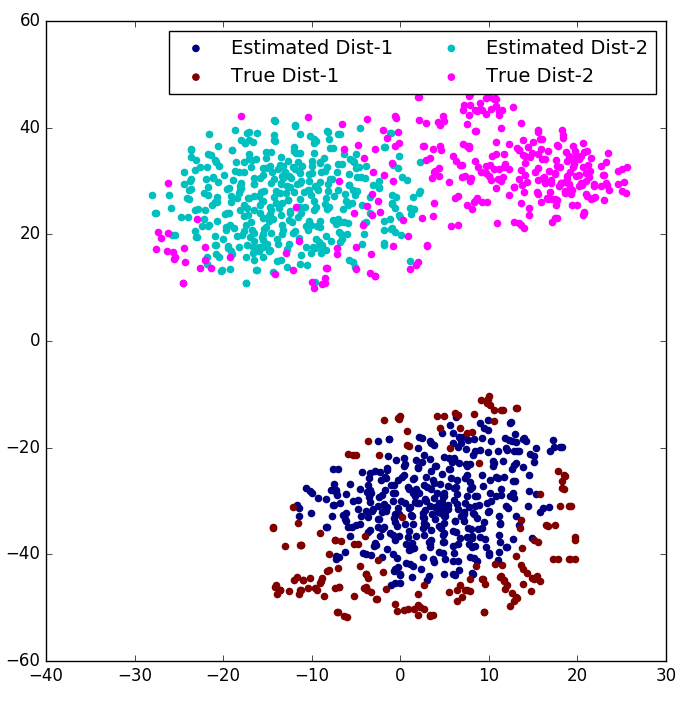}
\caption{t-SNE plot of estimated data distribution (using generated data) and true data distribution (using real data) for two of the unseen classes}
\label{fig:tsne}
\vspace{-1em}
\end{figure}
 the same classes. This is done by taking real and generated examples and embedding them into two dimensions using t-SNE. As shown in Fig.~\ref{fig:tsne} for two of the unseen classes, the empirical distributions of generated and real samples overlap significantly, corroborating our model's ability to generates samples that look like samples from the true distribution. 

\begin{figure}[!htbp]
\centering
\includegraphics[height=5cm,width=7cm]{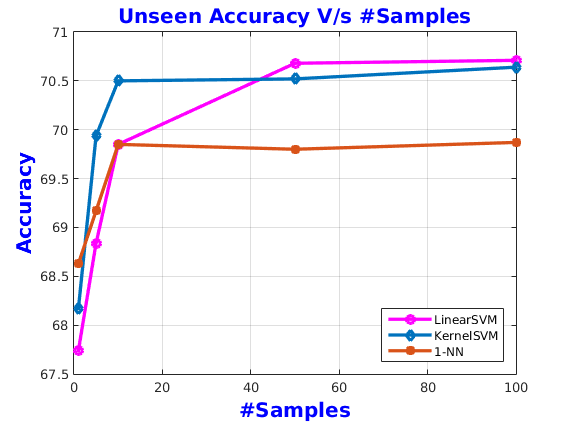}
\caption{Classification accuracies with varying \# of exemplars for AWA2 dataset.}
\label{fig:varygen}
\end{figure}

Finally, we also perform an experiment to assess how varying the number of the generated examples per class affects the classification accuracy. For this, we vary the number of generated examples per class in the range $[2,5,10,50,100]$, and use these examples in 3 off-the-shelf classifiers: linear SVM, kernel SVM, and nearest neighbors. As shown in Fig.~\ref{fig:varygen}, as expected, the classification accuracies increase with an increasing number of generated examples and it asymptotes fairly quickly, indicating that usually a small number of generated examples are sufficient to learn a fairly accurate classifier. 

\section{Discussion and Conclusion}
We have presented a robust generative framework to solve the generalized zero shot learning (GZSL) problem. Using a conditional VAE based architecture and augmenting it with discriminator-driven feedback mechanism enables our model to generate high-quality, class-specific exemplars from the unseen classes (and, if desired, also from seen classes). These exemplars can then be used in any classification model. This approach naturally helps us solve the GZSL problem since the learned classification model is not solely dependent on the labeled data from seen classes,  but also leverages synthesized examples from unseen classes. Our model can easily leverage unlabeled examples from seen and/or unseen classes and can therefore also operate in a semi-supervised setting. The model and the results presented here strongly demonstrate the effectiveness of learning continuous space models with significant power of generating exemplars representative of the true distribution. While we use a VAE style generative model for our case, extending it to adversarial training should enhance generated exemplar quality in terms of sharpness and realness as noted in \cite{progressivekarras}.  We also believe, the predictive power of the regressor can naturally improve performance in transductive ZSL settings~\cite{tran2015learning}, exploring which is a part of our future work. This can also be extended to online settings for few-shot learning where a small number of acquired labeled samples from the new classes can be used for improving the model.\\\\
\textbf{Acknowledgment:} Piyush Rai acknowledges support from Visvesvaraya Faculty Fellowship and a grant from Tower Research. Ashish acknowledges his travel support from ``Indian Summer School on Computer Vision, IIIT-H". Gundeep Arora acknowledges his travel support from Microsoft and Tower Research. Vinay Verma acknowledges support from Visvesvaraya fellowship and Research-I Foundation. 

{\small
\bibliographystyle{ieee}
\bibliography{egbib}
}
\end{document}